\documentclass{article}
\pdfoutput=1
\usepackage{spconf,amsmath,graphicx}
\usepackage[inkscapeformat=png]{svg}
\usepackage{rotating}
\usepackage{multirow}
\usepackage{url}
\usepackage{enumitem}
\setlist{nosep, leftmargin=14pt}

\usepackage{mwe} 

\title{PATHOLOGICAL PRIMITIVE SEGMENTATION BASED ON VISUAL FOUNDATION MODEL WITH ZERO-SHOT MASK GENERATION}
\name{Abu Bakor Hayat Arnob, Xiangxue Wang, Yiping Jiao, Xiao Gan, Wenlong Ming, and Jun Xu}
\address{Nanjing University of Information Science \& Technology, Nanjing, China}
\begin{document}
\ninept
\maketitle
\begin{abstract}
Medical image processing usually requires a model trained with carefully crafted datasets due to unique image characteristics and domain-specific challenges, especially in pathology. Primitive detection and segmentation in digitized tissue samples are essential for objective and automated diagnosis and prognosis of cancer. SAM (Segment Anything Model) has recently been developed to segment general objects from natural images with high accuracy, but it requires human prompts to generate masks. In this work, we present a novel approach that adapts pre-trained natural image encoders of SAM for detection-based region proposals. Regions proposed by a pre-trained encoder are sent to cascaded feature propagation layers for projection. Then, local semantic and global context is aggregated from multi-scale for bounding box localization and classification. Finally, the SAM decoder uses the identified bounding boxes as essential prompts to generate a comprehensive primitive segmentation map. The entire base framework, SAM, requires no additional training or fine-tuning but could produce an end-to-end result for two fundamental segmentation tasks in pathology. Our method compares with state-of-the-art models in F1 score for nuclei detection and binary/multiclass panoptic(bPQ/mPQ) and mask quality(dice) for segmentation quality on the PanNuke dataset while offering end-to-end efficiency. Our model also achieves remarkable Average Precision (+4.5\%) on the secondary dataset (HuBMAP Kidney) compared to Faster RCNN. The code is publicly available at \url{https://github.com/learner-codec/autoprom_sam}.
\end{abstract}

\section{Introduction}
\label{sec:intro}

The evolving landscape of medical image processing calls for automation to mitigate workforce shortages and escalating analysis costs. Machine learning, particularly deep learning, offers promise for efficient disease detection and diagnosis. However, acquiring an adequate, well-annotated mask for supervised training remains a challenge.

In our study, we propose a novel approach, an adaptation of the SAM, specifically tailored for segmentation tasks in digital pathology. SAM revolutionizes image segmentation by generating high-quality object masks from prompts, handling diverse inputs, and excelling in zero-shot scenarios. Our methodology involves generating bounding boxes from SAM's encoder layers and using them as prompts for SAM's pretrained lightweight mask decoder. This approach mitigates annotation complexity and time constraints. Our innovative technique aims to streamline the annotation process and ease the burden on pathologists by requiring significantly less time to draw bounding boxes around nuclei compared to the exhaustive annotation of nuclear boundaries for training while offering fine-grained segmentation masks during inference time.
Our contributions extend to three distinct areas:
\begin{itemize}
    \item An innovative feature extraction methodology is introduced to extract salient features from every layer of the transformer encoder, thus elevating overall performance.
    \item A distinctive architectural design is adopted that amalgamates the transformer encoder with a bottom-up configured convolutional neural network (CNN) decoder, enhancing the robustness of both detection and classification processes.
    \item An end-to-end network using SAM is presented to directly output classified objects in the form of bounding boxes and segmentation masks, eliminating the need for post-processing steps.
\end{itemize}

\section{RELATED WORKS}
\label{sec:related works}

The nuclei detection and segmentation domain primarily relies on U-Net-like \cite{Ronneberger2015} architectures featuring encoders and decoders. Encoders often adopt established structures like ResNet\cite{He2015}, transformers\cite{Zhai2021}, or as seen in Hover-Net\cite{Graham2019} and M-RCNN\cite{He2020}. Feature extraction strategies encompass direct or bottom-up approaches.

Recently, state-of-the-art methods[4] have improved segmentation by combining binary, distance, and nuclei-type maps, but are computationally expensive with multiple decoders. For natural images, DETR\cite{Carion2020} uses a transformer-encoder for bounding box prediction and utilizes hybrid encoders and transformer decoders for bounding box prediction. Nonetheless, DETR's 100 bounding box limitation poses issues with medical images, which can easily contain numerous nuclei instances. Adapting DETR to these images demands intricate and resource-intensive quantification. While CNN methods like \cite{He2020} display promise, they rely on both bounding boxes and segmentation masks for training, which requires a vast amount of manual annotation.
\begin{figure}
    \centering
    \includegraphics[width=8.5cm]{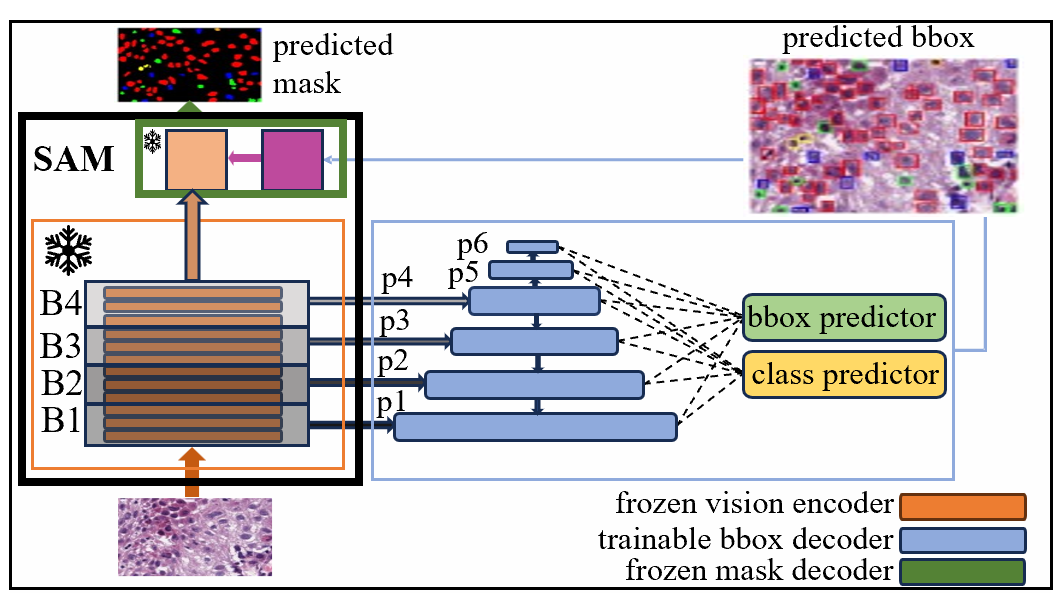}
    \caption{Proposed network architecture: transformer encoder (orange) for feature extraction, bounding box decoder (blue) for detection, and segmentation head (green) for precise segmentation.}
    \label{fig:enter-label}
\end{figure}

We present a novel method that can directly use domain-agnostic encoder features for all tasks while reducing fine-tuning overhead. This novel approach taps into diverse domains to alleviate the challenge of scarce annotated medical data, revealing a new dimension in pathological primitive detection.

\section{METHODOLOGY}
\label{sec:method}

Our foundational network, SAM \cite{Kirillov2023}, employs bounding boxes, points, text, or masks as prompts to generate segmentation masks, necessitating human intervention throughout the process. Moreover, its architecture lacks end-to-end capabilities, hindering streamlined analysis. We propose an extension to SAM (Fig. 1) that automatically generates segmentation prompts in the form of bounding boxes within the network and feeds them into the mask decoder. Our choice of bounding boxes as prompts stems from their superior performance observed in our initial experiments. In our approach, we completely freeze SAM's encoder and decoder, significantly reducing the trainable parameters(only 43.91 million) of the entire framework. In practice, only the bounding box predictor component is trained. This approach offers several key advantages: 1) it leverages the high-quality feature representation learned from natural images, reducing training time; 2) it seamlessly integrates bounding box detection and instance segmentation models, eliminating the need for pre- or post-processing.

\subsection{Frozen Transformer Encoder}
\label{ssec:frozen enc}

For our feature extraction backbone, we use the encoder from SAM\cite{Kirillov2023}, namely the lightweight SAM-B model. SAM has been trained on natural images and used for prompted generation. We propose that rich feature representation learned from natural images is sufficient for detection in the medical domain. As a result, we freeze the encoder entirely during training and use the embeddings from each layer for our decoder network.

\subsection{Bounding Box Decoder Network}
\label{ssec:frozen enc}

The decoder network in our proposed model exhibits a feature propagation process, which involves dividing the encoder into four distinct blocks, denoted as $\boldsymbol{B_i}$ where $i\in\{1,2,3,4\}$. The selection of these blocks is informed by the location of global attention applied in the encoder blocks. Furthermore, we introduce projection layers denoted as $\boldsymbol{p_{j\in\{1,2,3,4,5,6\}}}$ in our decoder, corresponding to each individual layer's feature from $\boldsymbol{B_{i\in\{1,2,3,4\}}}$ We utilize 6 projection layers because empirically it stabilizes our decoder network.

In total, our decoder consists of six projection layers, where we perform up-sampling on the first projection layer $\boldsymbol{p_1}$ and down-sampling in the projection layer $\boldsymbol{p_3}$ and $\boldsymbol{p_4}$, while keeping $\boldsymbol{p_2}$ unchanged. Each projection layer comprises a set of three convolutional layers with their corresponding up and down sampling and batch normalization, accompanied by a dense layer to increase the receptive field.

In the projection layer, feature aggregation occurs bottom-up, wherein the features from the three individual layers $\boldsymbol{l}$ are combined into one feature map. The dense layer includes a single convolutional operation, followed by a rectified linear unit (ReLU) activation function and batch normalization. After projection, we get the regression and classification logits $\boldsymbol{z_{j\in\{1,2,3,4,5,6\}}}$.

Additionally, we leverage skip connections from $\boldsymbol{p_1}$ to $\boldsymbol{p_4}$ to enhance information flow further and facilitate efficient gradient propagation during training. From $\boldsymbol{z_4}$, we generate additional feature maps, $\boldsymbol{z_5}$ and $\boldsymbol{z_6}$, using the projection layer $\boldsymbol{p_5}$ and $\boldsymbol{p_6}$. These feature maps have dimensions of $\boldsymbol{c\times h\in\frac{z_4}{2}\times w\in\frac{z_4}{2}}$ and $\boldsymbol{c\times h\in\frac{z_4}{4}\times w\in\frac{z_4}{4}}$ respectively. The comprehensive decoder architecture outlined above in Fig. 1 contributes significantly to our model's ability to capture and refine hierarchical features from the encoder. By aggregating features from multiple blocks and employing dense layers, we facilitate the integration of multi-scale information. The skip connections and feature maps also enhance the network's capacity to handle varying object sizes and scales, resulting in more accurate and context-aware detection.

In summary, we can define the whole process as follows:
\begin{equation}
\centering
{z'}_l = f_{enc}(x_i) \tag{1}
\end{equation} 

\begin{equation}
\centering
z_j = p_j ( z'_{({i'}, {i'}+3)\in {l}} ) \tag{2}
\end{equation} 
where $\boldsymbol{f_{enc}(.)}$ is the transformer encoder, $\boldsymbol{x_i}$ is the input image, $\boldsymbol{z'}$ is the extracted features from encoders, $\boldsymbol{i'}$ is the global attention index, $\boldsymbol{z_j}$ is the output from the projection layer $\boldsymbol{p_j}$, these projections are then fed into the regression and classification network.

\begin{equation}
\centering
y_{b,c} = f_{det}(z_j) \tag{3},
\end{equation}

\begin{equation}
\centering
y_{mask} = f_{seg}({z_{j^{'}}}, {y_{b,c}}) \tag{4},
\end{equation}

where $\boldsymbol{f_{det}(.)}$ is the bounding box regression and classification head, $\boldsymbol{f_{seg}(.)}$ is the mask segmentation head, $\boldsymbol{z_{j^{'}}}$ is the last layer's output from the encoder, $\boldsymbol{y_{b,c}}$ is the classified bounding box. The bounding box and classification decoder consist of three core components: anchor generation, bounding box regression, and classification. We adapt RetinaNet's\cite{Lin2017} anchor box generation technique, utilizing six labels instead of five, generated with aspect ratios [1:2, 1:1, 2:1].

\subsection{Frozen Mask Decoder}
\label{ssec:frozen mask}
The decoder component of the SAM\cite{Kirillov2023} is meticulously designed to deliver high-quality segmentation results, mainly focusing on producing refined segmentation masks near object boundaries. The decoder is trained with the encoder on natural images to achieve this. The decoder prioritizes high-quality segmentation, emphasizing refined masks at object boundaries. Building upon the encoder's already learned feature representation, we apply a frozen approach for the decoder. This strategic decision further limits the number of trainable parameters to only 43.91m in our network.

\subsection{Loss Function}
\label{ssec:loss}
Finally, we utilize the focal loss for training our network. Focal loss is a specialized loss function specifically designed for handling class imbalance in object detection and segmentation tasks. It addresses the common issue where the majority of the pixels or anchor boxes in medical images correspond to the background class, leading to an imbalanced distribution between foreground and background samples.

Mathematically, the focal loss is expressed as:
\begin{equation}
\centering
\text{Focal loss} = -\alpha(1-p_t)^{\gamma} \log(p_t) \tag{5},
\end{equation}

where ${\boldsymbol{p_t}}$ is the predicted probability of the ground truth class, $\boldsymbol{\alpha}$ and $\boldsymbol{\gamma}$ is the class balancing and modulating factor. In our practice we set $\boldsymbol{\alpha}$ to 0.5 and $\boldsymbol{\gamma}$ to 2.0.

\begin{figure}
    \centering
    \includegraphics[trim={0 5cm 0 0},clip, width=8.5cm]{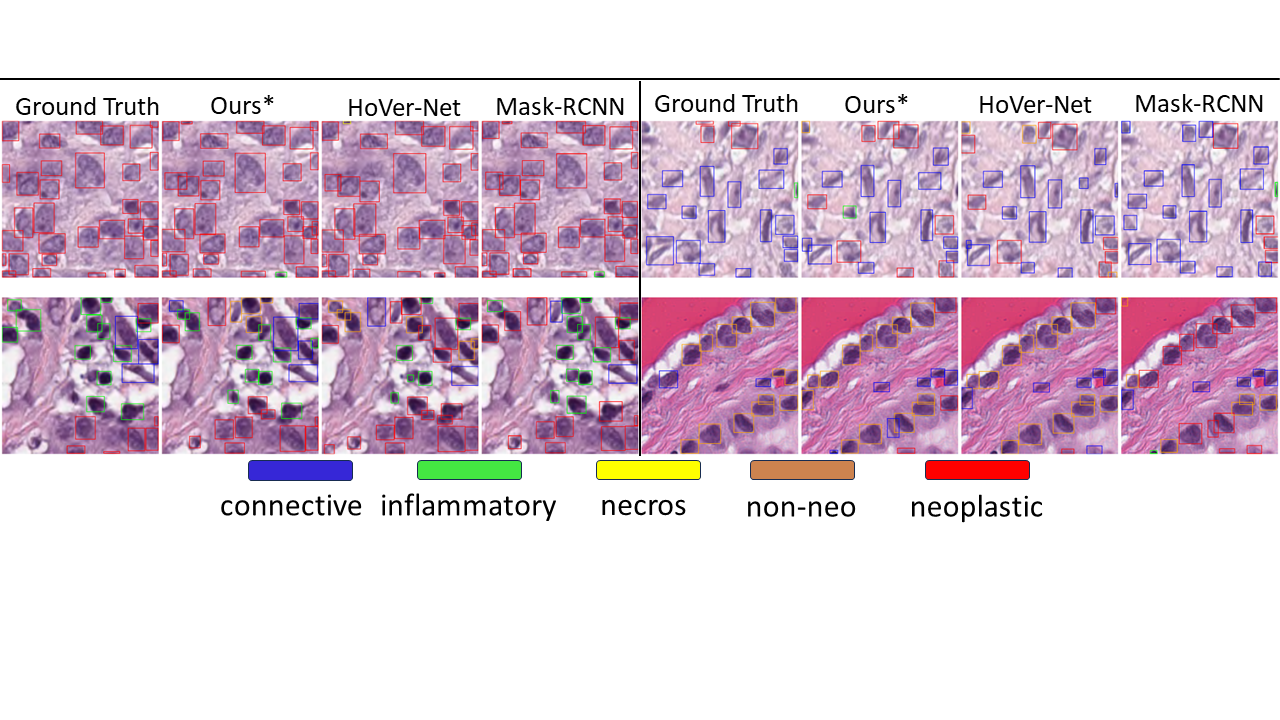}
    \caption{Bounding box detection quality and nuclei boundary visualization.}
    \label{fig:enter-label}
\end{figure}

\section{EXPERIMENTS}
\label{sec:experiments}

\subsection{Datasets}
\label{ssec:datasets}

For our investigation, we utilized the PanNuke\cite{Gamper2020} dataset as our primary dataset for training and evaluating our model. This dataset encompasses 7,904 images, each measuring 256 × 256 pixels, and contains 189,744 meticulously annotated nuclei. These nuclei span across 19 diverse tissue types. They are categorized into 5 distinct cell categories: connective, inflammatory, dead/necrotic, non-neo epithelial, and neoplastic, as illustrated in Fig. 2,  captured at a magnification of ×40. The cell images preserve a high level of resolution and intricate details of both nuclei and cells.

We also conducted a comprehensive evaluation utilizing a dataset obtained from HuBMAP\cite{hubmap2019human}, with a specific focus on glomeruli segmentation across 15 whole slide images. The dataset consists of 6694 patches extracted at a resolution of 2048x2048 pixels. Employing a three-fold cross-validation approach, 80\% of the data was dedicated to training and validation, while the remaining 20\% was exclusively reserved for testing.

\begin{table}[!h]
    \centering
        \begin{tabular}{c|c|c|c|p{1.2cm}}
        \hline
        Network & bPQ & mPQ & dice & trainable params$\downarrow$ \\ \hline
        Mask-RCNN & 0.5589 & 0.3688 & - & 172m \\
        Hover-Net & 0.7196 & $\boldsymbol{0.3726}$ & 0.8388 &54.74m \\
        Auto-Prom(Ours) & $\boldsymbol{0.7403}$ & 0.3618 & $\boldsymbol{0.8543}$ & 43.90m\\
        
        \hline
        \end{tabular}
    \caption{Comparison of bPQ, mPQ and dice scores along with trainable parameters.}
    \label{tab:table 1}
\end{table}

A noteworthy aspect of our primary dataset is the PanNuke dataset's inherent imbalance, with certain classes, namely dead and non-neo epithelial classes being highly underrepresented, exhibiting considerable underrepresentation. The dataset’s imbalance classes make it harder for models to learn and predict. 

\subsection{Experimental Setup}
\label{ssec:experimental_setup}
Our network undergoes 50 epochs with an initial learning rate of 3e-4, progressively reduced to 3e-7 through a reduction-on-plateau scheduler. Augmentations include rotation, flipping, noise injection, and color jittering. We implement a three-fold training approach with a holdout validation set on both datasets.

For our primary dataset, model evaluation is conducted on a 20\% test set randomly sampled from the three folds. This sampling ensures each fold contains an equal representation of the smallest class. Regarding the secondary dataset, evaluation is performed on a randomly sampled test set, as detailed in Section 4.1.

\begin{figure}
    \centering
    \includegraphics[clip, width=8.5cm]{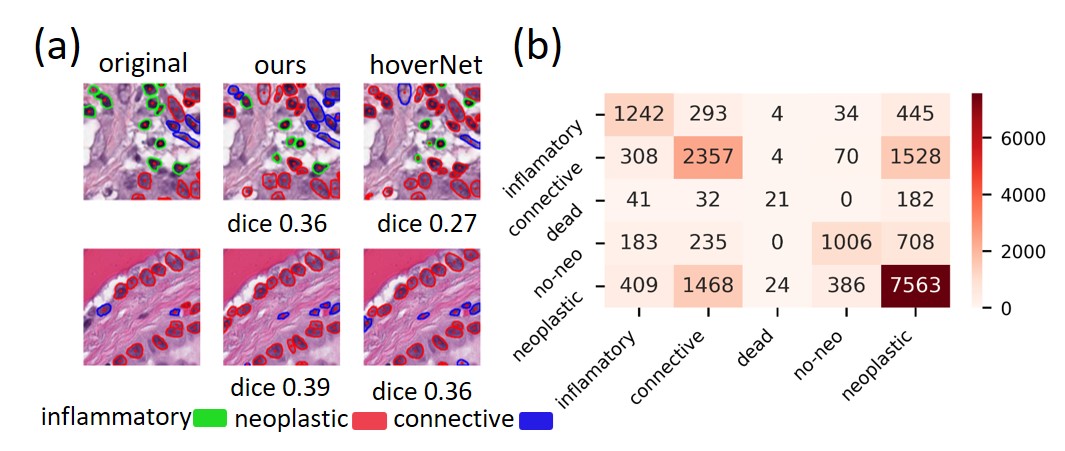}
    \caption{(a) Showcasing zero-shot(ours) and trained(hoverNet) multiclass mask quality and (b) Confusion matrix for multi-class classification results .}
    \label{fig:enter-label}
\end{figure}

\begin{table}[ht]
\centering
\begin{tabular}{|l|c|c|c|c|c|c|c|}
\hline
\multirow{2}{*}{Nuclei Labels} & \multicolumn{7}{c|}{Models} \\ \cline{2-8} 
 & \multicolumn{1}{l|}{} & \multicolumn{1}{l|}{M1} & \multicolumn{1}{l|}{M2} & \multicolumn{1}{l|}{M3} & \multicolumn{1}{l|}{M4} & \multicolumn{1}{l|}{M5} & \multicolumn{1}{l|}{Ours*} \\ \hline
& P & 0.40 & 0.49 & 0.55 & 0.59 & 0.58 & $\boldsymbol{0.70}$ \\ \cline{2-8} 
Neoplastic & R & 0.47 & 0.55 & 0.63 & 0.66 & 0.67 & $\boldsymbol{0.75}$ \\ \cline{2-8} 
 & F1 & 0.43 & 0.50 & 0.59 & 0.62 & 0.62 & $\boldsymbol{0.72}$ \\ \hline
& P & 0.27 & 0.38 & 0.52 & 0.63 & 0.54 & $\boldsymbol{0.71}$ \\ \cline{2-8} 
Non-Neo  & R & 0.31 & 0.33 & 0.52 & 0.54 & 0.54 & 0.54 \\ \cline{2-8} 
 & F1 & 0.29 & 0.35 & 0.52 & 0.58 & 0.56 & $\boldsymbol{0.61}$ \\ \hline
& P & 0.32 & 0.42 & 0.46 & 0.59 & 0.56 & $\boldsymbol{0.62}$  \\ \cline{2-8} 
Inflammatory  & R & 0.45 & 0.45 & 0.54 & 0.46 & 0.51 & $\boldsymbol{0.62}$ \\ \cline{2-8} 
 & F1 & 0.37 & 0.42 & 0.50 & 0.54 & 0.54 & $\boldsymbol{0.62}$ \\ \hline
& P & 0.34 & 0.42 & 0.42 & 0.50 & 0.52 & $\boldsymbol{0.53}$ \\ \cline{2-8} 
Connective & R & 0.38 & 0.37 & 0.43 & 0.45 & 0.47 & $\boldsymbol{0.54}$ \\ \cline{2-8} 
 & F1 & 0.36 & 0.39 & 0.42 & 0.47 & 0.49 & $\boldsymbol{0.54}$ \\ \hline
& P & 0.00 & 0.00 & 0.17 & 0.23 & 0.28 & $\boldsymbol{0.40}$ \\ \cline{2-8} 
Dead & R & 0.00 & 0.00 & 0.30 & 0.17 & $\boldsymbol{0.35}$ & 0.10 \\ \cline{2-8} 
 & F1 & 0.00 & 0.00 & 0.22 & 0.19 & $\boldsymbol{0.31}$ & 0.15 \\ \hline
\end{tabular}
\caption{Comparison of classification metrics with the baseline models. For the precision (P), recall (R) and F1 metric (F1), the centroid of the segmentation mask is used. The models are labeled as follows: M1: Det U-Net, M2: DIST, M3: Mask-RCNN, M4: Micro-Net, and M5: Hover-Net.}
\label{tab:second_table}
\end{table}

\subsection{Baseline and Evaluation metrics}
\label{ssec:experimental_setup}
In this study, we conducted three experiments. One of them is for
bounding box and mask prediction accuracy, which is done by comparing f1 accuracy. We also experiment with segmentation quality using panoptic quality\cite{kirillov2019panoptic} and dice score analysis on PanNuke dataset. Additionally, for our baseline comparison, we employ the baseline metrics provided by PanNuke\cite{Gamper2020} dataset organizers. All these metrics on the primary dataset for our network are acquired from the test split. In our comparative evaluation for mask quality analysis, we utilize Mask-RCNN and HoverNet as the baseline. The third experiment is conducted on HuBMAP dataset with an exclusive emphasis on utilizing bounding boxes on the secondary dataset. We choose Faster RCNN as our benchmark for comparing bounding box prediction performance(average precision). We quantified the Intersection over Union (IoU) overlap at 0.5 against the ground truth bounding boxes. If an IoU overlap with the ground truth is detected, the corresponding bounding box is associated with the respective bounding box.

\section{RESULTS AND DISCUSSIONS}
\label{sec:results}
For binary Panoptic Quality (PQ) analysis, the multiclass mask is converted into a binary mask to facilitate PQ metric generation. Our approach achieves superior performance over Mask-RCNN and Hover-Net in binary PQ and dice score detailed in Table 1, offering insights into the quality of masks. However Auto-Prom has a slightly poor performance in multi-class panoptic performance which is due to the under-representation of necrotic and non-neo epithelial class in PanNuke dataset and it is also noticeable by the recall score of these classes shown in Table 2. 

The results on the F1 accuracy are presented in Table 2. Our approach achieves state-of-the-art performance in multi-class classification across four distinct categories: neoplastic, inflammatory, and connective and non-neo epithelial.
Furthermore, we construct a confusion matrix based on our model's predictions. Fig. 3(b) visualizes this confusion matrix, elucidating the distribution and alignment of predictions within the different classification categories.

The analysis of the secondary dataset is elaborated in Table 3. Our approach demonstrates an Average Precision (AP) score of 79.89$\%$, representing a notable improvement of 4.5$\%$ over our baseline. The generated mask for the separate test set is visually depicted in  Fig. 4 for reference.
\begin{table}[!h]
    \centering
    \begin{tabular}{c|c}
        \hline
        Method & AP \\
        \hline
        Faster R-CNN (Res-101) & 75.33 \% \\
        Auto-Prom (ours) & $\boldsymbol{79.83}$ \% \\
        \hline
        \end{tabular}
    \caption{Bounding box quality analysis result on the secondary dataset. We use widely adopted average precision (AP) for the quality analysis between two models.}
    \label{tab:table 3}
\end{table}

In essence, our results underscore the prowess of our approach in attaining superior multiclass classification performance and elucidate the effectiveness of our methodology in the context of binary panoptic quality analysis and multi-class detection. The confusion matrix provides a comprehensive snapshot of prediction outcomes, contributing to a holistic understanding of our model's behaviour.

\section{CONCLUSION}
\label{sec:conclusion}

In conclusion, our approach delivers promising results and significantly enhances efficiency in pathological image analysis. By implementing our method, we achieve finer object boundaries without needing a ground truth segmentation dataset, relying solely on bounding boxes during training. This innovation could expedite the annotation process, allowing experts to outline bounding boxes and enabling automatic mask generation in SAM. Notably, our approach eliminates the need for prompts during inference, reducing human involvement.

\section{COMPLIANCE WITH ETHICAL STANDARDS}
\label{sec:ethical}

This research study was conducted retrospectively using human subject data made available in open access by \cite{Gamper2020},\cite{hubmap2019human}. Ethical approval was not required, as confirmed by the license attached with the open access data\cite{Gamper2020},\cite{hubmap2019human}.

\section{ACKNOWLEDGMENTS}
\label{sec:ackn}
This work was supported by the National Natural Science Foundation of China (Nos. 62171230, 92159301, 91959207, 62301265)

\begin{figure}
    \centering
    \includegraphics[width=8.5cm]{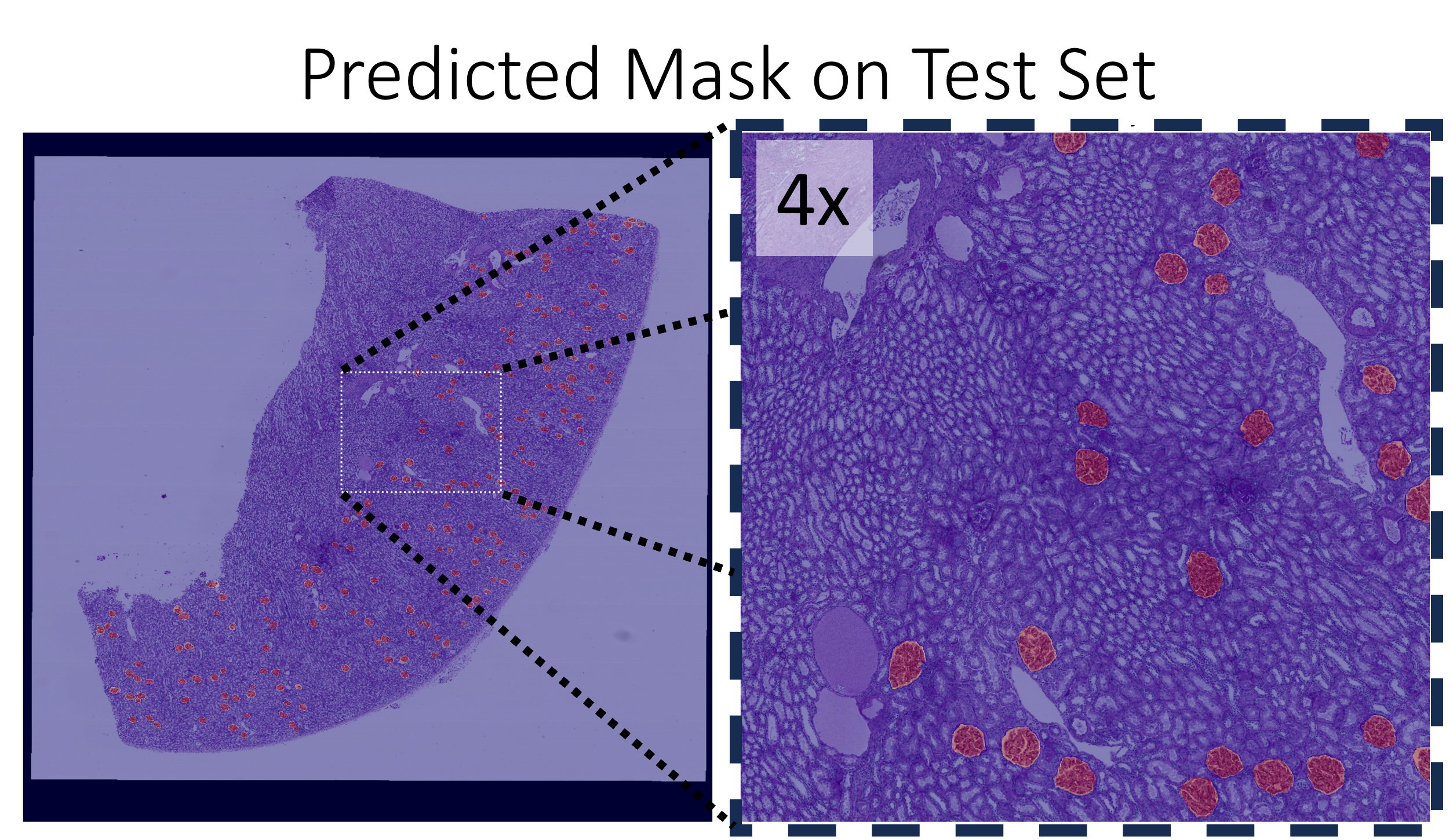}
    \caption{Predicted mask visualization on WSI from a test set of HuBMAP kidney segmentation dataset. The red regions correspond to our model’s prediction of glomeruli.}
    \label{fig:fig4}
\end{figure}

\bibliographystyle{IEEEbib}
\bibliography{strings,refs}

\end{document}